\documentclass[conference]{IEEEtran}
\IEEEoverridecommandlockouts

\usepackage{cite}
\usepackage{amsmath,amssymb,amsfonts}
\usepackage{graphicx}
\usepackage{textcomp}
\usepackage{xcolor}
\usepackage{booktabs}
\usepackage{url}
\usepackage{hyperref}
\usepackage{microtype}
\usepackage{multirow}
\usepackage{tikz}
\usetikzlibrary{arrows.meta, shapes.geometric, positioning, fit, backgrounds, calc, decorations.pathreplacing}

\hypersetup{colorlinks=true, linkcolor=blue, citecolor=blue, urlcolor=blue}
\begin{document}

\title{S23DR 2026: End-to-End 3D Wireframe Prediction\\
via DETR-Style Set Prediction with Contrastive Denoising}

\author{
\IEEEauthorblockN{Nitiz Khanal}
\IEEEauthorblockA{Pulchowk Campus, Lalitpur, Nepal}
}

\maketitle

\begin{abstract}
We present \textit{WireframeDETR}, our submission to the Structured Semantic
3D Reconstruction (S23DR) 2026 Challenge, which requires predicting a 3D
building wireframe from multi-view COLMAP point clouds.
Our method applies DETR~\cite{detr}-style set prediction directly to 3D point
clouds, producing wireframes as sets of edge coordinate pairs without any
intermediate vertex detection stage.
We introduce three technical contributions: (1)~\textbf{contrastive denoising
training} that stabilises noisy Hungarian matching in early epochs; (2)~a
\textbf{multi-scale encoder} that aggregates the last $K$ encoder layer
outputs via learned scalar weights; and (3)~\textbf{progressive auxiliary
loss weighting} that concentrates gradient signal on the decoder layers that
most benefit from it. Our model achieves a public test HSS of \textbf{0.575} (F1~=~0.664, IoU~=~0.516)
and a best validation HSS of \textbf{0.534} on the cleaned val split.
\end{abstract}

\section{Introduction}

The S23DR 2026 Challenge~\cite{s23dr2026} requires predicting the 3D
wireframe of a building rooftop — a set of 3D vertices and connecting edges —
from a sparse COLMAP~\cite{colmap} point cloud augmented with Gestalt and
ADE20K semantic labels and monocular depth estimates.
Performance is measured by the Hybrid Structure Score (HSS), the harmonic
mean of vertex F1 and edge volumetric IoU:
\begin{equation}
\mathrm{HSS} = \mathrm{harmonic\_mean}(F1(gt_v,\hat{v}),\,\mathrm{IoU}(gt_e,\hat{e})).
\label{eq:hss}
\end{equation}

We explored three directions before arriving at WireframeDETR.

\textbf{Path~A: Perceiver fine-tuning.}
The official learned baseline~\cite{baseline2026} is a Perceiver transformer
(val HSS~0.350). Resuming training with stronger augmentation caused
catastrophic forgetting: HSS fell below 0.26 within 20\,000 steps.

\textbf{Path~B: Two-stage PointNet pipeline.}
Motivated by the 2025 winning solution~\cite{skvrnjan2025}, we trained a
PointNet-style vertex detector (vertex F1~=~0.655, recall~=~0.927) followed
by an edge existence classifier (edge F1~=~0.289, precision~=~0.169).
The edge precision bottleneck proved difficult to close within the available
compute budget; we set this direction aside.

\textbf{Path~C: WireframeDETR (final approach).}
We reformulate wireframe prediction as direct edge-set regression over the
3D point cloud. Each predicted edge is a 6D coordinate pair
$(x_1,y_1,z_1,x_2,y_2,z_2)$; Hungarian matching assigns predictions to
ground-truth edges at training time. This avoids cascaded error propagation
and handles variable-cardinality wireframes without post-hoc NMS.

\section{Related Work}
\label{sec:related}

\textbf{3D wireframe and structured reconstruction.}
Classical methods recover building structure from point clouds via planar
primitive fitting and intersection.
PolyFit~\cite{polyfit} selects a minimal planar face set via integer
programming; City3D recovers LoD-2 models from airborne LiDAR using RANSAC
roof planes. Both require explicit intermediate representations and handcrafted
priors. The 2025 challenge winning solution~\cite{skvrnjan2025} uses two
PointNet-like networks: one refines 3D vertex candidates and a second classifies
per-pair edge existence, achieving HSS~=~0.43 on the private test set.

\textbf{DETR-style set prediction.}
DETR~\cite{detr} reformulates object detection as direct set prediction via
bipartite matching, eliminating NMS and hand-designed anchors.
DN-DETR~\cite{dndetr} identifies noisy Hungarian assignments in early training
as the chief convergence bottleneck and injects ground-truth-aligned denoising
queries to stabilise gradients.
DINO-DETR~\cite{dinodetr} further improves with contrastive denoising and
mixed query initialisation.
RoomFormer~\cite{roomformer} applies DETR-style set prediction to indoor floor
plan reconstruction, demonstrating that set-prediction transformers can replace
multi-stage graph pipelines for structured geometry.
We extend this paradigm to outdoor building wireframes operating on sparse 3D
point clouds.

\section{Dataset}
\label{sec:dataset}

The dataset (\texttt{usm3d/hoho22k\_2026\_trainval}) contains approximately
22\,000 building scenes. Each scene provides a COLMAP sparse point cloud with
per-point RGB colour, per-image Gestalt (73-class building-specific) and
ADE20K (107-class) segmentation maps, and monocular depth estimates from
Metric3Dv2~\cite{metric3dv2}. Ground-truth wireframes vary widely in complexity, from simple gable roofs
to highly articulated multi-ridge structures.

\begin{figure}[t]
\centering
\includegraphics[width=0.95\columnwidth]{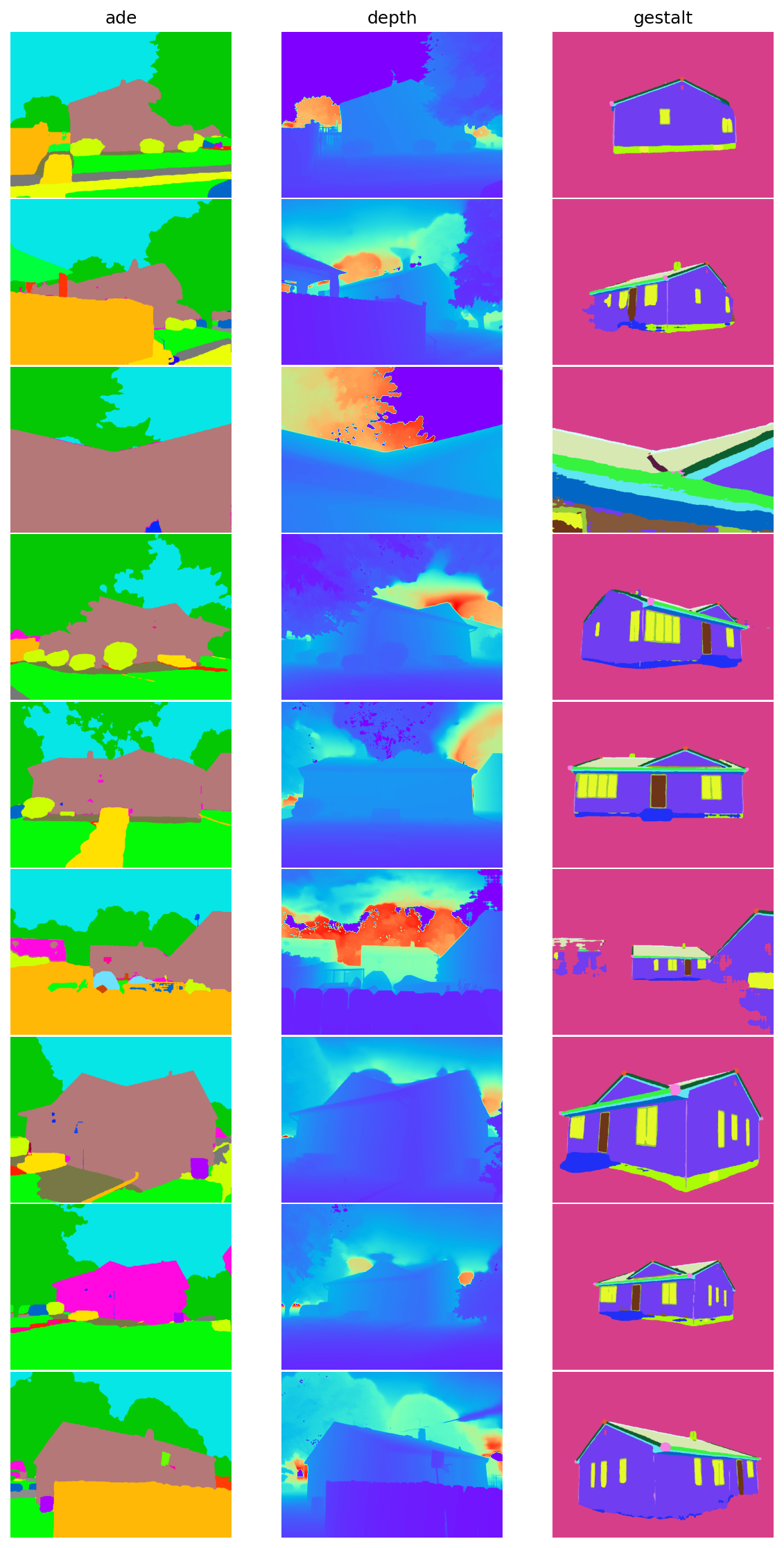}
\caption{Input modalities for a representative scene. Gestalt segmentation
labels directly encode vertex and edge classes of the roof structure, which
guides our sampling strategy.}
\label{fig:modalities}
\end{figure}

\textbf{Challenges.}
COLMAP reconstruction is sparse on textureless surfaces and frequently
incomplete. The dataset also contains scenes with inconsistent camera
parameters causing pose--annotation misalignment, where projected 3D vertices
do not align with their expected Gestalt class pixels. Both issues complicate
any approach that fuses the 3D point cloud with 2D segmentation information.
Figure~\ref{fig:modalities} illustrates the available input modalities.

\textbf{Point cloud preprocessing.}
Each 3D COLMAP point is projected into all visible views; the most common
Gestalt label across views is assigned to it~\cite{jastermark}. Points whose
dominant Gestalt label is \textit{unclassified} or \textit{unknown} are
discarded. The final per-point feature is the 3-channel RGB colour,
normalised to $[0,1]$.

\textbf{Gestalt-guided sampling.}
\label{sec:sampling}
Gestalt labels identify structurally significant point classes: vertex classes
(\textit{apex}, \textit{eave end point}, \textit{flashing end point}) and edge
classes (\textit{ridge}, \textit{rake}, \textit{eave}, \textit{hip}, etc.).
Following~\cite{jastermark}, we assign sampling weights 100, 10, and 1 to
vertex-class, edge-class, and background points respectively, then draw $N$
points proportional to these weights. This concentrates samples near structural
features without encoding semantic labels as model features. We draw
$N{=}4{,}096$ points during training and $N{=}7{,}168$ at inference.

\textbf{Coordinate normalisation.}
Point coordinates are normalised to $[0,1]^3$ per scene using the
$5$--$95$\,\% percentile range of each axis, providing a consistent scale
across scenes of varying physical size.

\textbf{Validation split.}
We evaluate on cleaned val: the official validation split filtered by a
curated list of scenes with severe pose--annotation misalignment, retaining
approximately 1\,150 clean scenes for evaluation.

\section{Method}
\label{sec:method}

\subsection{Architecture Overview}

\begin{figure}[t]
\centering
\begin{tikzpicture}[
  font=\scriptsize,
  node distance=0.38cm,
  enc/.style ={draw, rounded corners=2pt, fill=blue!10,   minimum width=3.2cm, minimum height=0.5cm, align=center, inner sep=3pt},
  mbox/.style={draw, rounded corners=2pt, fill=orange!15, minimum width=3.2cm, minimum height=0.5cm, align=center, inner sep=3pt},
  dec/.style ={draw, rounded corners=2pt, fill=purple!10, minimum width=3.2cm, minimum height=0.5cm, align=center, inner sep=3pt},
  cdn/.style ={draw, rounded corners=2pt, fill=orange!18, minimum width=2.6cm, minimum height=0.5cm, align=center, inner sep=3pt, dashed},
  obox/.style={draw, rounded corners=2pt, fill=green!12,  minimum width=1.4cm, minimum height=0.5cm, align=center, inner sep=3pt},
  lbl/.style ={font=\tiny, gray},
  arr/.style ={-{Stealth[length=4pt,width=3pt]}, semithick},
  skip/.style={-{Stealth[length=4pt,width=3pt]}, semithick, gray!55, dashed},
]

\node[enc]  (pts)  {Point Cloud\ \ $N\!\times\!3$ RGB};
\node[enc,  below=of pts]  (proj) {Linear Projection\ \ $3\!\to\!384$};
\node[enc,  below=of proj] (pe)   {SinCos 3D Positional Encoding};
\node[enc,  below=0.5cm of pe]    (e1)   {Enc Layer 1};
\node[enc,  below=of e1]   (e2)   {Enc Layer 2};
\node[enc,  below=of e2]   (e3)   {Enc Layer 3};
\node[enc,  below=of e3]   (e4)   {Enc Layer 4};
\node[mbox, below=0.5cm of e4]    (ms)   {Multi-Scale Aggregation\ \ $\sum_k w_k \mathbf{H}_k$};
\node[dec,  below=0.5cm of ms]    (dec)  {Transformer Decoder\ \ 5 Layers};

\node[obox, below=0.5cm of dec, xshift=-0.85cm] (cls) {Class\\Head};
\node[obox, below=0.5cm of dec, xshift= 0.85cm] (crd) {Coord\\Head 6D};

\node[enc,  right=1.1cm of ms]    (qry) {$Q{=}128$\\Queries};
\node[cdn,  above=0.4cm of qry]   (cdn) {CDN Module\\(train only)};

\draw[arr] (pts)  -- (proj);
\draw[arr] (proj) -- (pe);
\draw[arr] (pe)   -- (e1)  node[midway,right,lbl]{$N\!\times\!384$};
\draw[arr] (e1)   -- (e2);
\draw[arr] (e2)   -- (e3);
\draw[arr] (e3)   -- (e4);
\draw[arr] (e4)   -- (ms)  node[midway,right,lbl]{last 3 layers};
\draw[arr] (ms)   -- (dec) node[midway,right,lbl]{memory};
\draw[arr] (dec)  -- (cls);
\draw[arr] (dec)  -- (crd);

\draw[skip] (e2.east) -- ++(0.3,0) |- (ms.east);
\draw[skip] (e3.east) -- ++(0.2,0) |- (ms.east);

\draw[arr,dashed] (cdn) -- (qry) node[midway,right,lbl]{GT noise};
\draw[arr] (qry.south) |- (dec.east);

\end{tikzpicture}
\caption{WireframeDETR architecture. RGB features are projected to $d{=}384$,
enriched with sinusoidal positional encoding, and processed by a 4-layer
Transformer encoder. Multi-scale aggregation combines the last $K{=}3$ layer
outputs via learned softmax weights (grey taps). At training time the CDN
module injects denoising queries with known GT assignments; at inference only
the $Q{=}128$ learned queries are used.}
\label{fig:arch}
\end{figure}
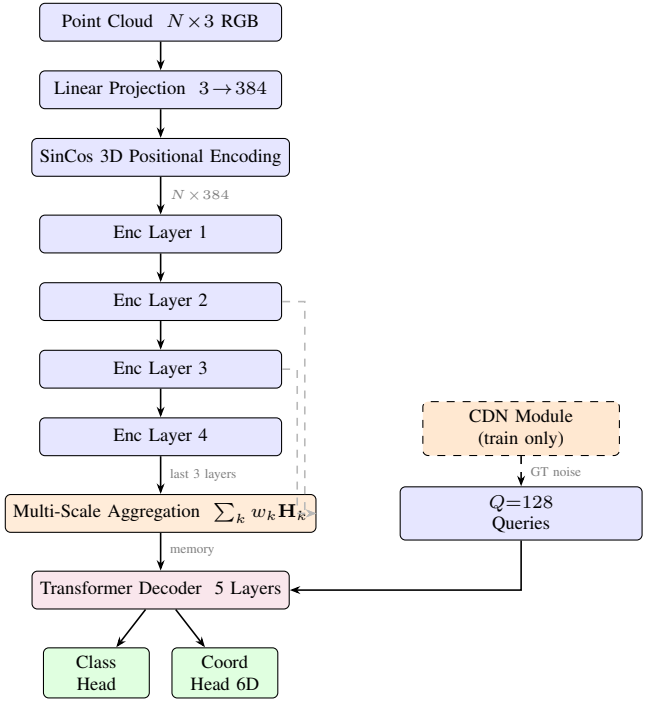

Figure~\ref{fig:arch} illustrates \textit{WireframeDETR}. A linear projection
maps the 3-dimensional RGB features to embedding dimension $d=384$.
A 3D sinusoidal positional encoding (SinCos3DPE) splits $d$ evenly across
XYZ axes and is added to each point token. A 4-layer multi-scale Transformer
encoder processes all $N$ point tokens. A 5-layer Transformer decoder with
$Q=128$ learned edge queries produces the final predictions.

At training time, Hungarian matching assigns predictions to ground-truth edges
minimising a combined cost of L1 coordinate distance and negative
classification probability. The training loss has two terms: cross-entropy
(weight~1.0) and L1 coordinate regression (weight~5.0). Auxiliary outputs
from each intermediate decoder layer are supervised with progressively
increasing weight (Section~\ref{sec:progaux}).

The model has approximately 22.7\,M parameters.

\subsection{Contrastive Denoising Training}
\label{sec:cdn}

Standard DETR training suffers from unstable Hungarian assignments in early
epochs, where the same query is matched to different GT targets across
iterations, producing contradictory gradients~\cite{dndetr}.
We address this with contrastive denoising (CDN) adapted from
DN-DETR~\cite{dndetr}.

\begin{figure}[t]
\centering
\begin{tikzpicture}[font=\small, scale=0.85]
  \def\Q{3}  
  \def\D{2}  

  \foreach \i in {0,...,4} {
    \foreach \j in {0,...,4} {
      \pgfmathparse{(\i<\Q && \j>=\Q) || (\i>=\Q && \j<\Q) ? 1 : 0}
      \ifnum\pgfmathresult=1
        \fill[red!40] (\j*0.55, -\i*0.55) rectangle ++(0.55,-0.55);
      \else
        \fill[green!20] (\j*0.55, -\i*0.55) rectangle ++(0.55,-0.55);
      \fi
      \draw[gray!50] (\j*0.55,-\i*0.55) rectangle ++(0.55,-0.55);
    }
  }
  \node[above] at (0.825,0.05)  {\tiny Learned $Q$};
  \node[above] at (2.475,0.05)  {\tiny CDN};
  \node[left]  at (0,-0.275)    {\tiny $Q$};
  \node[left]  at (0,-1.375)    {\tiny CDN};
  \node[below right, green!60!black] at (0.05,-2.8) {\tiny \checkmark\ can attend};
  \node[below right, red!70!black]   at (1.60,-2.8) {\tiny $\times$\ masked};
\end{tikzpicture}
\caption{CDN attention mask. Learned and denoising queries attend only within
their own group, keeping gradient paths independent.}
\label{fig:attnmask}
\end{figure}
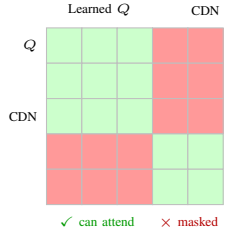

At each training step, for each of the $M$ ground-truth edges we generate
$G{=}5$ \textit{positive} denoising queries by perturbing GT coordinates with
noise $\epsilon \sim \mathcal{U}[-\lambda_+, \lambda_+]^6$,
$\lambda_+=0.4$, and $G{=}5$ \textit{negative} queries with larger noise
$\lambda_- = 0.8$. Positive queries are supervised against their originating
GT edge; negative queries are supervised as background. An attention mask
(Figure~\ref{fig:attnmask}) prevents cross-attention between denoising and
learned queries, preserving independence.

The decoder sequence length at training time is $Q + 2GM$ (varying per scene
as $M$ differs). The CDN loss uses the same coefficients as the main loss.
During training, the CDN coordinate loss decreases from $\approx$2.07 at
epoch~0 to $\approx$1.25 at epoch~15, confirming that the denoising path
converges steadily alongside the main queries.

\subsection{Multi-Scale Encoder Memory}

A standard Transformer encoder passes only the final layer's output to the
decoder, discarding intermediate representations. We collect all encoder layer
outputs $\{\mathbf{H}_1, \ldots, \mathbf{H}_L\}$ and compute a learned
weighted sum of the last $K{=}3$ outputs:
\begin{equation}
  \mathbf{M} = \sum_{k=1}^{K}
    \frac{e^{w_k}}{\sum_{j=1}^{K} e^{w_j}}\,
    \mathbf{H}_{L-K+k},
  \label{eq:multiscale}
\end{equation}
where $w_1, w_2, w_3 \in \mathbb{R}$ are learned scalar weights initialised
to~0. This gives the decoder access to both local geometry (earlier layers)
and abstract structural features (final layer) at a cost of three extra
parameters.

\subsection{Progressive Auxiliary Loss Weighting}
\label{sec:progaux}

Auxiliary outputs after intermediate decoder layers carry steadily improving
prediction quality; we reflect this by weighting the $i$-th auxiliary output
($i=0,\ldots,N_\mathrm{aux}-1$) as:
\begin{equation}
  w_i^\mathrm{aux} = 0.5 + 0.5\,\frac{i+1}{N_\mathrm{aux}},
  \label{eq:progaux}
\end{equation}
linearly increasing from 0.6 (first auxiliary) to 1.0 (pre-final layer).

\subsection{Post-Processing and Inference}

Predicted edges are filtered by classification score ($>0.9$). The resulting
vertex set is de-duplicated by iteratively merging vertices within 0.5\,m,
placing each merged vertex at the intersection of incident line
directions~\cite{jastermark}. We adopt this post-processing from publicly
available code~\cite{jastermark}.

Optionally, four Y-axis rotations ($0^\circ, 90^\circ, 180^\circ, 270^\circ$)
can be applied at inference time; predictions from each rotation are
un-rotated and merged via the same vertex deduplication step. WireframeDETR
treats points as an unordered set and is therefore compatible with rotational
augmentation at inference. This is disabled by default as its contribution
was not isolated in ablation.

\subsection{Training Setup}

We train with AdamW ($\mathrm{lr}{=}10^{-4}$, weight decay $10^{-4}$), a
one-cycle LR schedule, batch size~14, and FP16 mixed precision on a single
NVIDIA A100 80\,GB GPU for 200 epochs ($\approx$27\,hours). Data augmentation:
random Y-axis rotation and random horizontal flip. We disable
\texttt{torch.compile} because CDN's variable decoder sequence length
($Q + 2GM$) triggers repeated Triton kernel recompilation. Gradient $\ell_2$
norm is clipped at 0.1 to stabilise CDN training.

\section{Experiments}
\label{sec:experiments}

\subsection{Main Results}

Table~\ref{tab:results} summarises validation HSS across our development
progression. WireframeDETR with plain 3-channel RGB features and Gestalt-guided
sampling achieves a public test HSS of \textbf{0.575} and a best
validation HSS of 0.534, substantially exceeding the official Perceiver
baseline (0.350).

\begin{table}[t]
\centering
\caption{Results across development stages.
$^\dagger$Public leaderboard (test split); Vert.\ F1 and Edge IoU from the
leaderboard \texttt{corner\_f1} and \texttt{edge\_iou} metrics.
$^\ddagger$Cleaned validation split.
Path~B leaderboard scores reflect the full two-stage system; internally
the vertex detector achieved F1~=~0.655 and the edge classifier F1~=~0.289.}
\label{tab:results}
\begin{tabular}{lcccc}
\toprule
Approach & Split & Vert.\ F1 & Edge IoU & HSS \\
\midrule
Perceiver baseline~\cite{baseline2026}   & Val$^\ddagger$   & ---   & ---   & 0.350 \\
\quad + fine-tuning (Path~A)             & Val$^\ddagger$   & ---   & ---   & $<$0.26 \\
PointNet two-stage (Path~B)              & Test$^\dagger$  & 0.497 & 0.409 & 0.442 \\
\midrule
WireframeDETR (ours)                     & Val$^\ddagger$   & 0.603 & 0.471 & 0.534 \\
WireframeDETR (ours, \textbf{best})      & Test$^\dagger$  & \textbf{0.664} & \textbf{0.516} & \textbf{0.575} \\
\bottomrule
\end{tabular}
\end{table}

\subsection{CDN Training Dynamics}

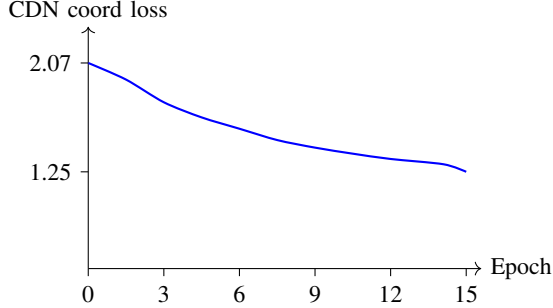
\begin{figure}[t]
\centering
\begin{tikzpicture}[font=\small]
\begin{scope}
  \draw[->] (0,0) -- (5.2,0) node[right] {Epoch};
  \draw[->] (0,0) -- (0,3.2) node[above] {CDN coord loss};

  \draw[blue, thick] plot[smooth] coordinates {
    (0,   2.72)   
    (0.5, 2.5)
    (1.0, 2.2)
    (1.5, 2.0)
    (2.0, 1.85)
    (2.5, 1.70)
    (3.0, 1.60)
    (3.5, 1.52)
    (4.0, 1.45)
    (4.7, 1.38)
    (5.0, 1.28)
  };

  \foreach \y/\label in {1.28/1.25, 2.72/2.07} {
    \draw (0,\y) -- (-0.1,\y) node[left] {\label};
  }
  \foreach \x/\label in {0/0, 1.0/3, 2.0/6, 3.0/9, 4.0/12, 5.0/15} {
    \draw (\x,0) -- (\x,-0.1) node[below] {\label};
  }
\end{scope}
\end{tikzpicture}
\caption{CDN coordinate loss over the first 15 epochs. Steady convergence
confirms the denoising path provides clean gradient signal from epoch~0.}
\label{fig:cdncurve}
\end{figure}

Figure~\ref{fig:cdncurve} shows the CDN coordinate loss over the first 15
epochs. The loss decreases steadily from 2.07 to 1.25, confirming that
denoising queries — which have fixed GT assignments bypassing noisy Hungarian
matching — converge reliably from the first iteration.
CDN classification loss falls from 0.170 to 0.137 over the same window.
Both are consistently lower than the corresponding main loss terms,
as expected when queries receive cleaner gradient signal.

\subsection{Auxiliary Loss Ordering}

At all observed training epochs the auxiliary losses satisfy
$\ell_\mathrm{aux\_0} < \ell_\mathrm{aux\_1} < \ell_\mathrm{aux\_2} <
\ell_\mathrm{aux\_3} < \ell_\mathrm{main}$, confirming that later decoder
layers produce progressively better predictions and that progressive weighting
does not cause any layer to under-perform its predecessors.

\subsection{Comparison with Baselines}

On the public leaderboard, WireframeDETR (HSS~=~0.575) outperforms both the
handcrafted baseline (HSS~=~0.391~\cite{s23dr2026}) and the official learned
Perceiver baseline (HSS~=~0.474~\cite{baseline2026}) by a substantial margin.

\subsection{Limitations}

Training is $\approx$2$\times$ slower than the Perceiver baseline
($\approx$8\,min/epoch vs.\ 4\,min/epoch). CDN's variable-length decoder
sequence precludes \texttt{torch.compile}, adding $\approx$30\% overhead.
TTA (4 rotations) multiplies inference time by $4\times$ but requires no
additional training; it is disabled by default in our submission.

\section{Conclusion}
\label{sec:conclusion}

We presented WireframeDETR, a DETR-based end-to-end model for 3D building
wireframe prediction that directly regresses edge coordinate pairs from a
sparse RGB point cloud. Our three contributions — contrastive denoising training, multi-scale encoder
memory, and progressive auxiliary loss weighting — combine to achieve a public
test HSS of \textbf{0.575} and a best validation HSS of 0.534 on the S23DR
2026 cleaned val split. Post-processing routines are
adopted from~\cite{jastermark}. Future work includes adding vertex feature
heads for feature-guided vertex merging and exploring joint vertex--edge
prediction with contrastive vertex losses.



\begin{thebibliography}{99}

\bibitem{s23dr2026}
J. Langerman, D. Mishkin, and Y. Huang,
``S23DR: Structured Semantic 3D Reconstruction Challenge 2026,''
\url{https://huggingface.co/spaces/usm3d/S23DR2026}, 2026.

\bibitem{baseline2026}
J. Langerman,
``S23DR 2026 learned baseline (Perceiver),''
\url{https://huggingface.co/usm3d/learned-baseline-2026}, 2026.

\bibitem{jastermark}
jastermark,
``S23DR2026: Open-source EdgeDETRPE implementation,''
\url{https://huggingface.co/jastermark/S23DR2026}, 2026.

\bibitem{skvrnjan2025}
J. Skvrna and L. Neumann,
``Structured Semantic 3D Reconstruction (S23DR) Challenge 2025 -- Winning
Solution,''
\emph{arXiv preprint}, 2025.

\bibitem{detr}
N. Carion, F. Massa, G. Synnaeve, N. Usunier, A. Kirillov, and S. Zagoruyko,
``End-to-end object detection with transformers,''
in \emph{Proc. ECCV}, 2020, pp.~213--229.

\bibitem{dndetr}
F. Li, H. Zhang, L. Liu, J. Guo, L. Ni, and L. Zhang,
``DN-DETR: Accelerate DETR training by introducing query denoising,''
in \emph{Proc. CVPR}, 2022, pp.~13619--13627.

\bibitem{dinodetr}
H. Zhang, F. Li, S. Liu, L. Zhang, H. Su, J. Zhu, L. Ni, and H.-Y. Shum,
``DINO: DETR with improved denoising anchor boxes for end-to-end object
detection,''
in \emph{Proc. ICLR}, 2023.

\bibitem{colmap}
J. L. Sch\"{o}nberger and J.-M. Frahm,
``Structure-from-motion revisited,''
in \emph{Proc. CVPR}, 2016, pp.~4104--4113.

\bibitem{metric3dv2}
M. Hu et al.,
``Metric3D v2: A versatile monocular geometric foundation model for
zero-shot metric depth and surface normal estimation,''
\emph{IEEE TPAMI}, 2024.

\bibitem{polyfit}
L. Nan and P. Labatut,
``PolyFit: Polygonal surface reconstruction from point clouds,''
in \emph{Proc. ICCV}, 2017, pp.~2353--2361.

\bibitem{roomformer}
J. Yue et al.,
``RoomFormer: Encoding floor plans as sequences of rooms,''
in \emph{Proc. CVPR}, 2023, pp.~2121--2130.

\end{thebibliography}
\end{document}